\documentclass[lettersize,journal]{IEEEtran}
\usepackage{amsmath,amsfonts}
\usepackage{algorithmic}
\usepackage{algorithm}
\usepackage{array}
\usepackage{textcomp}
\usepackage{stfloats}
\usepackage{url}
\usepackage{verbatim}
\usepackage{graphicx}
\usepackage{cite}
\usepackage{url}
\usepackage{booktabs} 
\usepackage{subfigure}

\usepackage{mathtools}
\usepackage{amsthm}
\newtheorem{theorem}{Theorem}[section]

\newtheorem{lemma}[theorem]{Lemma}

\begin{document}

\title{Lift What You Can: Green Online Learning with Heterogeneous Ensembles}

\author{
    \IEEEauthorblockN{Kirsten K\"obschall\IEEEauthorrefmark{1}\IEEEauthorrefmark{3}, Sebastian Buschj\"ager\IEEEauthorrefmark{2}, Raphael Fischer\IEEEauthorrefmark{2}, Lisa Hartung\IEEEauthorrefmark{1}, Stefan Kramer\IEEEauthorrefmark{1}}
    \IEEEauthorblockA{\\\IEEEauthorrefmark{1}Johannes Gutenberg University Mainz, Mainz, Rhineland-Palatinate, Germany}
    \IEEEauthorblockA{\\\IEEEauthorrefmark{2}Technical University of Dortmund, Dortmund, North Rhine-Westphalia, Germany}
    \IEEEauthorblockA{\\\IEEEauthorrefmark{3}\textit{Corresponding author}; koebschall@uni-mainz.de}
}
%




\maketitle

\begin{abstract}
Ensemble methods for stream mining necessitate managing multiple models and updating them as data distributions evolve.
Considering the calls for more sustainability, established methods are however not sufficiently considerate of ensemble members' computational expenses and instead overly focus on predictive capabilities.
To address these challenges and enable green online learning, we propose \textit{\textbf{he}te\textbf{r}ogeneous \textbf{o}nline en\textbf{s}embles} (HEROS).
For every training step, HEROS chooses a subset of models from a pool of models initialized with diverse hyperparameter choices under resource constraints to train. 
We introduce a Markov decision process to theoretically capture the trade-offs between predictive performance and sustainability constraints.
Based on this framework, we present different policies for choosing which models to train on incoming data. Most notably, we propose the novel $\zeta$-policy, which focuses on training near-optimal models at reduced costs.
Using a stochastic model, we theoretically prove that our $\zeta$-policy achieves near optimal performance while using fewer resources compared to the best performing policy.
In our experiments across 11 benchmark datasets, we find empiric evidence that our $\zeta$-policy is a strong contribution to the state-of-the-art, demonstrating highly accurate performance, in some cases even outperforming competitors, and simultaneously being much more resource-friendly.
\end{abstract}

\begin{IEEEkeywords}
Data stream learning, Resource-awareness, Evolving data streams.
\end{IEEEkeywords}

\section{Introduction}
Environmental sustainability in the context of any technological progress is enshrined in the United Nations' \emph{Agenda 2030} and is also elemental in the first wave of artificial intelligence (AI) regulations \cite{EU_AI_Act_2024}.
Despite the imperative need for resource-awareness and sustainable development in machine learning (ML) \cite{schwartz_green_2020,van_wynsberghe_sustainable_2021}, the ``bigger-is-better paradigm'' \cite{varoquaux_hype_2024} is however still observable in the field \cite{luccioni_power_2024}.
This also holds true for the domain of stream mining, where challenges like drift reaction and online processing necessitate sophisticated methods and good hyperparameter tuning \cite{VELOSO202175}.
For such data, online ensemble learning is popular because of its adaptability and good predictive performance, however, it does not sufficiently acknowledge the importance of resource efficiency.
While resource efficiency in terms of processing speed has long been considered in stream mining research, the explicit trade-offs between predictive performance and resource-constraints have not been investigated much until recently. 

\begin{figure}[t]
    \centering
    \includegraphics[width=0.95\linewidth]{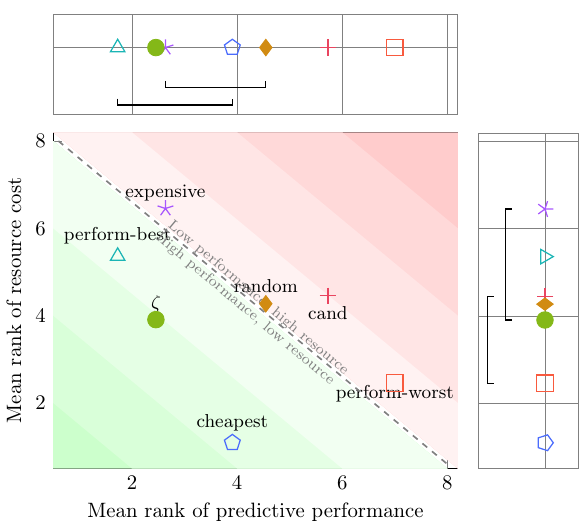}
    \caption{Trade-off performance and resource costs in kWh of HEROS under different policies in terms of the mean ranks over 11 data streams and 3 random repetitions. Our $\zeta$-policy has a high predictive performance and a low resource cost while training. 
    The critical differences, determined using the Wilcoxon signed-rank test with $95\%$ confidence level and adjusted with Holm's method, are positioned along the axis -- above for the performance metric and to the right for resources.}
    \label{fig:trade-off}
\end{figure}
In this paper, we propose heterogeneous online ensemble (HEROS) to enable resource-aware (i.e., green) online learning.
Our method makes informed decisions when selecting and training a subset of models in the heterogeneous pool.
For every single training step, it chooses the models under resource constraints for efficiently enhancing the ensemble's overall predictive capability.
HEROS should be understood as a theoretical framework for capturing online learning under resource constraints.
This, for example, encompasses the recent CAND approach \cite{DBLP:conf/ijcnn/GunasekaraGPB22} (for tuning hyperparameters on data streams) and easily allows developing additional policies.
In particular, we also propose the novel $\zeta$-policy and theoretically show that it has comparable performance to updating the best-performing models but uses much fewer resources (cf. Fig. \ref{fig:trade-off}).
In short, our contributions are the following:
\begin{enumerate}
    \item We introduce HEROS, a novel framework for training heterogeneous ensembles in a resource-constraint online setting. HEROS enables a customizable and transparent control of resources while online training. 
    \item We present a formulation of online learning in terms of a Markov decision process (MDP), allowing us to unify various selection policies for training the ensemble members. We also offer a novel resource-constrained policy, for which we present a stochastic proof that analyzes the predictive performance and resource cost at the training selection step.
    \item We discuss experiments showing that our proposed algorithm achieves better performance than established selection policies (Figure \ref{fig:trade-off}), while substantially saving resources during training. 
\end{enumerate}

This work is structured as follows: Important related work is discussed in Section \ref{sec:related_work}. 
Subsequently, Section \ref{sec:algorithm} presents the problem formulation, the HEROS framework, and our new $\zeta$ policy. 
Section \ref{sec:theory} introduces a stochastic model and gives a stochastic proof on the comparison of three policies.
The experimental results are presented in Section \ref{sec:experiments} and in Section \ref{sec:conclusion} we conclude our work.

\section{Related Work}\label{sec:related_work}
Efficient processing and adapting to changes in data through ensemble methods are two of the key cornerstones in stream data mining \cite{DBLP:journals/corr/abs-2004-05785, DBLP:journals/inffus/KrawczykMGSW17, DBLP:journals/sigkdd/GomesRBBG19}. 
Efficiency in streaming mainly focuses on how quickly data can be processed while keeping resource use low.
The Hoeffding tree algorithm by Domingos \cite{DBLP:conf/kdd/DomingosH00} is a classic example, designed to build decision trees fast and with limited memory.
This algorithm has been fine-tuned over the years to further boost its speed and cut down on resource needs, while being tested in various scenarios \cite{DBLP:conf/pkdd/HolmesKP05, DBLP:conf/kdd/ManapragadaWS18, DBLP:journals/ijdsa/Garcia-MartinLG21}.
Similarly, other classical algorithms such as Online Naive Bayes \cite{DBLP:conf/bigdataconf/BahriMB18} and the online SVM \cite{DBLP:conf/icpr/Liu016} have been made more efficient over the years.
Handling concept drift in data streams often involves using ensemble methods to keep models effective as data changes.
Initially, this meant tweaking batch ensemble techniques like Bagging and Boosting for online use \cite{DBLP:conf/aistats/OzaR01}.
Recent efforts have pushed this idea further, exploring the theoretical underpinnings of online boosting \cite{DBLP:conf/nips/BeygelzimerHKL15, DBLP:conf/icml/ChenLL12} and adapting complex strategies like Random Forests for real-time data through approaches like Adaptive Random Forests \cite{DBLP:journals/ml/GomesBRBEPHA17} and Streaming Random Patches \cite{DBLP:conf/icdm/GomesRB19}.
Multi-armed bandits (MAB) also need to be named in the context of iterative learning, as they use training information to evaluate the actions based on reward \cite{SuttonBarto-RL}, which makes them capable of addressing problems over data streams.

Alongside these developments, there however is a growing need for making ML and data stream mining more sustainable, resource-aware, and transparent, as also evidenced by AI regulatories \cite{EU_AI_Act_2024} and several related articles \cite{luccioni_power_2024,fischer_towards_2024}.
They call for novel methods that explicitly consider user preferences regarding the trade-offs between predictive capabilities, resource consumption, and model interpretability \cite{fischer_metaqure,fischer_xpcr}, however, few papers put this into practice for online learning.
One recent example are Soft Hoeffding Trees \cite{10.1007/978-3-031-78977-9_11}, combining gradient-based learning with Hoeffding Trees to efficiently train differentiable trees that also allow for post-hoc explainability.
Paying close attention to resource consumption, the Shrub Ensemble method \cite{DBLP:conf/aaai/BuschjagerHM22} combines ensemble pruning via proximal gradient descent with online learning in order to offer a fixed memory consumption while keeping speed and accuracy up.

Several ensemble methods use only a single model at prediction time, such as Weighted Majority Algorithm \cite{littlestoneWarmuth1994}, Cost-sensitive Tree \cite{pmlr-v28-xu13}, and Dynamic Classifier Selection \cite{dynamicclassifierselection2018}, which all aim to retain ensemble benefits while reducing inference complexity.
Online learning frameworks such as MOA are extended to use measurement tools like \emph{Intel}'s Running Average Power Limit (RAPL) for tracking energy consumption \cite{DBLP:conf/bracis/OnukiMB23}.
Finally, and arguably most relevant to our paper, Gunasekara et al. recently proposed the CAND method for online learning of multiple neural networks while paying close attention to the update costs of each network \cite{DBLP:conf/ijcnn/GunasekaraGPB22}.
As updating the complete model pool can become costly, the authors propose to focus on updating the currently best performing models and a small random selection to still be able to follow concept drifts. 
This paper builds upon this idea, generalizes the ad-hoc training selection policy \cite{DBLP:conf/ijcnn/GunasekaraGPB22}, and presents a more principled and sophisticated approach.

\section{Methodology}

\label{sec:algorithm}
\begin{figure}[t]
    \centering
    \includegraphics[width=\linewidth]{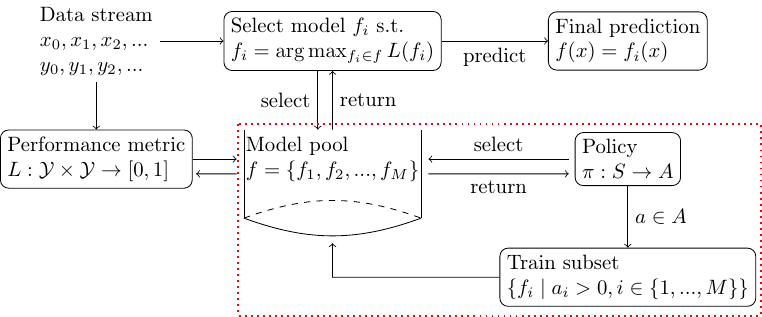}
    \caption{Schematic overview of HEROS to handle the incoming data stream, involving the model selection for prediction $f(x)$ and the associated model selection for training using a policy $\pi$ within the model pool $f$.}
    \label{fig:schema}
\end{figure}

We propose the HEROS framework for training an ensemble of models over a data stream under resource constraints. At its core, HEROS is an ensemble of models each initialized with different hyperparameters, where only a subset of models is trained as data arrives to save costly resources. To do so, HEROS uses a \emph{policy} to determine which models should be updated in each training step. 
However, each model can provide feedback on the incoming data stream during training, so we can view resource-constrained training as a Markov decision process (MDP), which is explained in detail next.
Then, after introducing some intuitive policies and the CAND approach \cite{DBLP:conf/ijcnn/GunasekaraGPB22}, we propose the novel $\zeta$-policy.
Its performance in comparison to other policies will be theoretically and practically analyzed in Section \ref{sec:theory} and \ref{sec:experiments}, respectively.

\subsection{Problem Formulation}
We consider a supervised learning setting where $\mathfrak{S} = \{(x_t,y_t)\mid t\in\{0,\dots\}\}$ is an open-ended sequence of feature vectors $x_t \in \mathcal X$ and labels $y_t\in \mathcal Y$.
For simplicity, we focus on classification problems with $C$ classes (i.e.,  $\mathcal{Y} = \{1,2,\dots,C\}$) and $d$-dimensional feature vectors (i.e., $\mathcal X \subseteq \mathbb R^d$), although our method trivially extends to regression problems and other types of features (e.g., text or images).
Let the function $f_i$ be a model for the iteratively observed data from $\mathfrak{S}$, with $f_i: \mathcal{X} \rightarrow \mathcal{Y}$, $i \in \{1, \dotsc, M\}$, and we further define the model pool $f$ as $f:= \{ f_1, f_2, \dots, f_M \}$.
For any batch of observed data $(x_t, y_t)$, the complete pool is evaluated to assess the individual model performance w.r.t. some normalized performance metric $L(f_i(x_t), y_t)$ that is \emph{maximized} (e.g., accuracy), with $L: \mathcal{Y} \times \mathcal{Y} \rightarrow [0, 1]$. During training and whenever labels are available during the stream, we record the performance of each model.
For improved readability, we remove the dependency on $x_t$ and $y_t$ in the following, and as such use $L(f_i)$ to refer to the recorded performance of any model, while the overall ensemble performance is defined as $\mathcal{L}(f) = \max_{f_i \in f} L(f_i)$.
In addition, let $\gamma = \left(\gamma_1, \gamma_2, \dots, \gamma_M \right)\in \mathbb{R}^M$ be a vector representing the resource consumption, where $\gamma_i$, is the required cost to perform a single training step with $f_i$. W.l.o.g. $0 \leq \gamma_i \leq 1$. 

To predict the class label for the current instance $x$, $f_i$ is selected such that $f_i = \arg\max_{f_i\in f}L(f_i)$ and used as prediction $f(x)$ of the ensemble $f$ (cf. Fig. \ref{fig:schema}).
In this paper, we focus on the selection of models for training using a policy $\pi$ given a state space $S$ to determine the action space $A$.
Continuously improving $f$ requires finding a set of $k$ models ($k \leq M$, $k \in \mathbb{N}_{>0}$) to train at each update step,  with preferably small resource consumption and good quality on the current data.
Selecting these $k$ models over a stream of incoming data can be formulated as an MDP $(S, A, \mathcal{T}, R)$ \cite{PUTERMAN1990331} which is defined as follows:
\begin{itemize}
    \item \textbf{State Space:} $S=\{ \left( f_{1,r_1}, \dots f_{M,r_M} \right) \mid r_i \in \mathbb{N}_{\geq 0}, i \in \{1, \dotsc, M\} \}$ is the state space, where $r_i$ denotes the already invested resource cost to train $f_i$. We denote a state at time $t>0$ as $s_t\in S$.
    Note $f_{i,r_i}$ is not unique under $r_i$, since if $r_i$ was invested (trained) at a different time point, the $f_{i,r_i}$'s may differ.
    \item \textbf{Action Space:} $A=\{ a \mid a \in \{ 0,1 \}^M, \sum_{i=1}^M a_i \leq k \}$ is the set of actions denoting which $k$ models to train per step, in order to reduce resource consumption (if $k<M$).
    As such, $a_i = 1$ when resources $\gamma_i$ are spent to train $f_{i,r_i}$, and $a_i = 0$ if $\gamma_i$ are saved and $f_{i,r_i}$ remains unchanged. 
    \item \textbf{Transition Function:} $\mathcal{T}(s_t,a)$ is the transition function for each entry $i$ in the state tuple, being defined as
    \begin{align}
        \mathcal{T}(s_t,a)_{(i)}=
        \begin{cases}
            f_{i,r_i} & \text{if}\: a_i = 0 \\
            f_{i,r_i + \gamma_i} & \text{if}\: a_i = 1
        \end{cases}
    \end{align}
    \item \textbf{Reward:} $R(s_t, a)$ is the reward function based on the performance metric $L$ and model training costs $\gamma$, and is defined as
        \begin{align}
            R(s_t, a) = \sum_{i=1}^{M} ( L(f_i) + (1-\gamma_i) ) \cdot a_i.
        \end{align}
    As such, training is either rewarded due to low costs ($1-\gamma_i$) or high performance ($L(f_i)$)\footnote{The reward $R$ could also be implemented with a weighted approach in the form $L(f_i) + w(1-\gamma_i))$, where $w\in \mathbb R$ is a user preference. However, this would lead to the challenge of Pareto optimality \cite{fischer_xpcr}, which is very expensive to solve optimally.
    Instead, we developed a more efficient policy to guide this trade-off via a more intuitive,  user-controllable threshold.}. We will revisit the specific use of $R$ in Section \ref{subsec:zeta-policy}.
\end{itemize}
With this formalization, training HEROS corresponds to the proposed MDP based on a policy $\pi$ that guides decision-making by providing actions $a$ at any point in time. 
Instead of solving this MDP with reinforcement learning, which tends to be slow in online scenarios \cite{10.5555/3618408.3618475}, we use theory-guided (policy in Section \ref{subsec:zeta-policy}, analysis in Section \ref{sec:theory}) heuristic policies.

\subsection{Pragmatic Policies}\label{subsec:allpolicies}
Choosing a subset of $k$ models to train, or in other words, determining a specific action $a$ given $\pi$ and $s_t$, should maximize the performance and minimize the resource costs of a training step.
Similar to \cite{fischer_metaqure}, this represents a multi-objective optimization problem which we encode in the reward $R$, i.e. our goal is to determine $\arg\max_{a\in A} R(s_t, a)$ at any state $s_t$.
Exhaustively testing actions $a\in A$ by maximizing $R$ corresponds to the Knapsack problem, which is known to be NP-complete, and therefore requires a more sophisticated approach that ideally also incorporates user preferences.
Before we introduce such an approach for addressing this multi-objective optimization problem, our novel $\zeta$-policy, let us consider a few straightforward policies:
First, the \texttt{random} policy selects $k$ models uniform randomly without considering performance or costs, i.e., $a$ has $k$ random non-zero entries.
The \texttt{perform-best} policy chooses $k$ individual models with the best performance $\max_{f_i \in f}L(f_i)$, while \texttt{perform-worst} chooses them with $\min_{f_i \in f}L(f_i)$.
With \texttt{cheapest}, the $k$ distinct models with lowest resource cost are chosen (i.e., $f_i$ is selected if $i=\arg\min_{i\in\{1, \dotsc, M\}}\gamma_i$) and \texttt{expensive} complementarily picks $k$ models with $i=\arg\max_{i\in\{1, \dotsc, M\}}\gamma_i$.
As explained earlier, the \texttt{CAND} policy \cite{DBLP:conf/ijcnn/GunasekaraGPB22} is a mixture of these approaches, selecting $\lfloor \frac{k}{2} \rfloor$ models via the best performance and the other half at random.

\subsection{$\zeta$-Policy}
\label{subsec:zeta-policy}

\begin{algorithm}[tb]
   \caption{$\zeta$-Policy}
   \label{alg:zeta-policy}
\begin{algorithmic}[1]
   \STATE {\bfseries Input:} Model pool $f$, number of models to train $k$, $\zeta$, performance metric $L$
   \STATE Initialize action $a:= 0^M$, $J=\{1, \dotsc, M\}$
   \FOR{$1$ {\bfseries to} $k$}
        \STATE Set $j = \arg\max_{j\in J}L(f_j)$
        \STATE Set $i=j$
        \FOR{{\bfseries each} model $f_l \in f$ with $L(f_l) \geq (1-\zeta) L(f_j)$}
            \IF{$\gamma_l < \gamma_{i}$ and $a_{i}=0$} 
                \STATE $i = l$
            \ENDIF
        \ENDFOR
        \STATE $a_{i} = 1$, $J = J\setminus \{i\}$
   \ENDFOR
   \STATE {\bfseries return} $a$
\end{algorithmic}
\end{algorithm}

In order to balance predictive performance and resource consumption in more sophisticated ways, we introduce the new $\zeta$-policy described in Algorithm \ref{alg:zeta-policy}.
Instead of following an ad-hoc approach or expensively testing all options, this policy represents a greedy approach to maximize $R$ while considering user priorities via the threshold $\zeta \in [0, 1]$.
Following the intention of $R$, the $\zeta$-policy selects a model $f_i$ with the lowest resource costs $\gamma_i$, which has an estimated predictive performance not worse than $1-\zeta$ of the best not-yet-chosen model $f_j$.
For $k\leq M$ distinct models, we determine the model $f_j$, with the highest predictive performance relative to $L$, and within the area $1-\zeta$ around $L(f_j)$, we investigate whether there exists a model $f_i$ with strict lower resource costs. If this is the case, it is selected by setting the respective $a_i$ to $1$.
The performance of each model (regardless of whether the model is trained) is updated given the current instance, which leads to a transit in the reward state, so that the order of $f_1, \dotsc, f_M$ given their performance changes.
By evaluating all models at each step, we aim to maximize the long-term expected reward by training models that perform effectively in state $s_t$.
This is particularly important for responding to concept drifts. 
So far, this policy prioritizes the best current states by selecting actions based on exploitation. To explore the remaining state space, however, a random combination of $k$ models is selected using the $\epsilon$-greedy strategy with a probability of $\epsilon$. Using $\epsilon >0$ serves as an optional optimization for the $\zeta$-policy.

\section{Theoretical Analysis} \label{sec:theory}
In this section, we analyze the asymptotic behavior of the average performance and resource costs of the selected models from three policies: $\zeta$, \texttt{CAND}, and \texttt{perform-best}.
We prove that the average performance of the model selection converges in probability under $\zeta$-policy to a higher value than for \texttt{CAND} policy for $\zeta$ small.  Moreover, it is at most $\zeta$ worse than the \texttt{perform-best} policy.
Furthermore, we show that the average resource consumption under $\zeta$-policy is smaller than under \texttt{CAND} and \texttt{perform-best}. 
First, we introduce a stochastic model to analyze the distribution and relationships of the states and policies introduced in Section \ref{sec:algorithm}.

\subsection{A Stochastic Model}
\label{subsec:stochastic_model}
Let $X:=\{X_1, X_2, \dots, X_M\}$ be a set of independent and identically distributed random variables, where $X_i=L(f_i)$, $i\in \{1, \dots, M\}$, $f_i \in f$, and $X_i \sim \text{Beta}(\alpha,\beta)$, $\alpha, \beta > 0$.
Moreover, let $\{ \gamma_1, \gamma_2, \dots, \gamma_M\}$  be a set of uniform distributed random variables over $\left\{ \frac{1}{M}, \frac{2}{M}, \dots, \frac{M}{M} \right\}$ describing the resource consumption\footnote{The arguments carry over verbatim to the setting where the density of $X_i$ and the density of $\gamma_i$ has support $[0,1]$.}. Note, $X_i$ and $\gamma_i$, are independent of each other\footnote{The assumption of independence of $X$ and $\gamma$ is a simplification. However, as long as the joint distribution of $(X,\gamma)$ has a strictly positive density in $[0,1]^2$, i.e. $P(X,\gamma) > 0, \forall (X,\gamma) \in [0,1]^2$, this will effectively not change the qualitative results of the theoretical analysis, but makes the analysis less transparent.}. 
In the theoretical analysis of the policies, we take $M$ to infinity and analyze the asymptotic behavior of their performance and resource consumption under the stochastic model described above. Taking $M$ to infinity implies that we are examining a pool encompassing models with every possible configuration of hyperparameters. 
Firstly, we analyze the two policies individually.

Let
\begin{align*}
X_{(M)}\geq X_{(M-1)}\geq \dots \geq X_{(1)}
\end{align*}
denote the order statistics of $X$.
The \texttt{CAND} policy selects half of the $k$ models according to the order statistics of $X$ and the other half at random.
We denote the first half (best-performing) as a tuple of performance and resource cost as follows 
\begin{align*}
    (X_{(M)}, \gamma_{1,b}^C), (X_{(M-1)}, \gamma_{2,b}^C), \dotsc, (X_{(M-\lfloor\frac{k}{2}\rfloor)}, \gamma_{\lfloor\frac{k}{2}\rfloor,b}^C).
\end{align*}
The performance and resources of the randomly chosen models are denoted by
\begin{align*}
    (X_{1,r}^C, \gamma_{1,r}^C), (X_{2,r}^C, \gamma_{2,r}^C), \dotsc, (X_{\lceil\frac{k}{2}\rceil,r}^C, \gamma_{\lceil\frac{k}{2}\rceil,r}^C).
\end{align*}
The overall performance under \texttt{CAND} policy is
\begin{align*}
    X^C_k = \sum_{i=1}^{\lfloor\frac{k}{2}\rfloor} X_{(M-i-1)} + \sum_{i=1}^{\lceil\frac{k}{2}\rceil} X_{i,r}
\end{align*}
and the overall resource consumption is
\begin{align*}
    \gamma^C_k = \sum_{i=1}^{\lfloor\frac{k}{2}\rfloor} \gamma_{i,b} + \sum_{i=1}^{\lceil\frac{k}{2}\rceil} \gamma_{i,r}.
\end{align*}
Due to readability, we omit the detailed proofs here and refer interested readers to Appendix.
\begin{lemma}\label{lemma:cand}
    The asymptotic behavior of the average of $\frac{1}{k}X_k^C$ and $\frac{1}{k}\gamma_k^C$ converges in probability to $\frac{1}{2}+\frac{1}{2} \mathbb{E}(X)$ and $\mathbb{E}(\gamma)$, respectively, when first $M\to\infty$ and then $k\to \infty$.
\end{lemma}

Now looking at our $\zeta$ policy, we denote the $k$ models selected by it (cf. Alg. \ref{alg:zeta-policy}) as a tuple of performance and resource cost
\begin{align*}
    (X_1^{\zeta}, \gamma_1^{\zeta}), (X_2^{\zeta}, \gamma_2^{\zeta}), \dotsc, (X_k^{\zeta}, \gamma_k^{\zeta}).
\end{align*}
We consider the case using the $\zeta$-policy with probability $(1-\epsilon)$, where the $k$ models with the smallest resources are chosen from the set
$
  \{i\vert  X_i^\zeta \geq (1-\zeta) X_{(M)} \}.
$
As we are taking $M$ large in our theoretical analysis, $X_{(M)}$ converges almost surely to one. Hence, a valid approximation is to choose the $k$ models with the smallest resources from the set 
\begin{align*}
  \{i\mid  X_i^\zeta \geq (1-\zeta) \}.
\end{align*}
For $M$ sufficiently large, this set contains more than $k$ elements\footnote{By the strong law of large numbers
$     \lim_{M\to \infty} \frac{1}{M} |\{i\mid  X_i^\zeta \geq (1-\zeta) \}| =\mathbb{P}(X_i \geq 1-\zeta) \quad \mbox{almost surely}$. }.
\begin{lemma}\label{lemma:zeta}
    The asymptotic behavior with probability $1-\epsilon$ of the average of $\frac{1}{k}X^\zeta$ and $\frac{1}{k}\gamma^\zeta$ under $\zeta$-policy converges in probability to $\mathbb{E}(X\mid X\geq 1-\zeta)$ resp. $0$ as first $M\to\infty$ and then $k\to\infty$.
\end{lemma}
Now that we have estimated the asymptotic behavior of the performances and resources for both policies, we can compare the results.

\begin{theorem}\label{theorem:performance-cand}
    The average performance in state $s_t$ with probability $1-\epsilon$, achieved by applying the $\zeta$-policy, is higher than that achieved by applying \texttt{CAND} policy as $M \to \infty $ and then $k\to \infty$ if  
    $$
     \frac{1}{2} (1-\mathbb{E}(X)) > \zeta.    $$
     If $X$ is Beta($\alpha,\beta$)- distributed, this holds if 
     $\zeta < \frac{1}{2} (1- \frac{\alpha}{\alpha + \beta} )$.
\end{theorem}
Thus, for small $\zeta$, the average performance of the selected models under $\zeta$-policy is in probability higher than under \texttt{CAND}. We now analyze the resource consumption for training in state $s_t$.
\begin{theorem}\label{theorem:ressource_cand}
    The average resource consumption in state $s_t$ with probability $1-\epsilon$, achieved by applying the $\zeta$-policy, is lower than that achieved by applying \texttt{CAND} policy for $M\to\infty$ and then $k\to\infty$.
\end{theorem}
In other words, with probability $1-\epsilon$, i.e. if we do not choose our action at random, the resource consumption under $\zeta$-policy is asymptotically lower than under \texttt{CAND}.
Finally, we investigate the relation of $\zeta$-policy and \texttt{perform-best}. 
\texttt{perform-best} policy selects $k$ models with the highest performance, i.e. $X_{(M)}, X_{(M-1)}, \dotsc, X_{(M-k)}$ denoted by 
\begin{align*}
    (X_{(M)}, \gamma_1^{pb}), (X_{(M-1)}, \gamma_2^{pb}), \dotsc, (X_{(M-k)}, \gamma_k^{pb}).
\end{align*}
The overall performance under \texttt{perform-best} policy is $X^{pb} = \sum_{i=1}^k X_{(M-i-1)}$ and the overall resource cost is $\gamma^{pb} = \sum_{i=1}^k \gamma_i^{pb}$.
In particular, given that we want to maximize the predictive performance, we can make the following statement:
\begin{theorem}
\label{theorem:perform-best-perf}
    The average performance during training in state $s_t$ with probability $1-\epsilon$, achieved by applying the $\zeta$-policy, is at most $\zeta$ worse than that achieved by \texttt{perform-best} policy for large $M$ and $k$, where $M>k$.
\end{theorem}
Thus, we are at most $\zeta$ worse than under $\zeta$-policy, however:
\begin{theorem}\label{theorem:ressource_performbest}
    The average resource consumption during training at state $s_t$ with probability $1-\epsilon$, achieved by applying the $\zeta$-policy, is lower than that achieved by applying \texttt{perform-best} policy for large $M$ and $k$, where $M>k$.
\end{theorem}
To summarize, the resource consumption under $\zeta$-policy is asymptotically lower than under \texttt{CAND} and \texttt{perform-best} while leveraging models with higher or at most $\zeta$ worse predictive performance, respectively.

\section{Experiments}
\label{sec:experiments}
In the experiments, we begin by analyzing the distribution of $X_i$ from Section \ref{sec:theory}. We show that the Beta distribution approximates the $X_i$ accurately.
Second, we compare the proposed HEROS w.r.t. resource consumption and predictive performance using the $\zeta$-policy to all other policies introduced in Section \ref{subsec:allpolicies}, which is \texttt{random}, \texttt{perform-best}, \texttt{perform-worst}, \texttt{cheapest}, \texttt{expensive}, and \texttt{CAND} policy.
To clarify the effect of drift adaptation, we analyze gradual and abrupt drift reaction of HEROS.
Finally, we compare the predictive performance of the state-of-the-art ensemble methods, adaptive random forest (ARF), streaming random patches (SRP), and Shrub ensembles (Shrub) from CapyMOA \cite{capymoa_2024} to our proposed HEROS.
\begin{table}[t]

    \caption{Dataset properties. The data type is denoted as follows: (\textbf{D})rifts, (\textbf{R})eal-world, (\textbf{S})ynthetic.}
    \centering
    \resizebox{\linewidth}{!}{
    \begin{tabular}{l  l l  l l l l}
        Name & Type & Instances & Features & \# Classes & \multicolumn{2}{c}{Class distribution} \\
        & &&&& max(\%) & min(\%) \\
        \toprule
        airlines & R &  539382 & 7 & 2 & 55.46 & 44.54 \\
        electricity & R & 45310 & 8 & 2 & 57.55 & 42.45 \\
        WISDM & R & 5417 & 45 & 6 & 38.43 & 4.53 \\
        covtype & R & 581010 & 54 & 7 & 48.76 & 0.47 \\
        nomao & R & 34464 & 118 & 2 & 71.44 & 28.56 \\
        \midrule
        AGR$_a$ & DS & 1000000 & 9 & 2 & 53.68 & 46.31 \\
        AGR$_g$ & DS & 1000000 & 9 & 2 & 53.68 & 46.31 \\
        RBF$_f$ & DS & 1000000 & 10 & 5 & 26.11 & 9.27 \\
        RBF$_m$ & DS & 1000000 & 10 & 5 & 26.11 & 9.27 \\
        LED$_a$ & DS & 1000000 & 24& 10 & 10.08 & 9.94 \\
        LED$_g$ & DS & 1000000 & 24& 10 & 10.08 & 9.94 \\
        \bottomrule
    \end{tabular}
    }
    \label{table:datasets}
\end{table}
The datasets include real-world and synthetic streams as classification benchmarks used in the  literature \cite{DBLP:conf/ijcnn/GunasekaraGPB22, DBLP:conf/icdm/GomesRB19}.
An overview of our 11 investigated datasets is given in Table \ref{table:datasets}, and the parameters for the introduced concept drifts as obtained from the synthetic stream generator can be found in our repository.
For evaluating HEROS, we trained respective ensembles ($M=50$ members) with policies introduced in Section \ref{subsec:allpolicies} and $\zeta$ with different hyperparameters $\zeta\in \{ 0.01, 0.05 \}$ and $\epsilon \in \{ 0.1, 0.2 \}$. 
Enabling a fair comparison, we also allow exploration for \texttt{perform-best}, \texttt{perform-worst}, \texttt{cheapest}, and \texttt{expensive} by applying an $\epsilon$-greedy approach with $\epsilon = 0.1$.
As base learners in the ensemble, we evaluate multi-layer perceptrons (MLP) and Hoeffding trees (HT).
The MLPs are configured to use Adam and SGD optimizer, all combinations of five learning rates $\{5\cdot 10^{-1}, 5\cdot 10^{-2}, 5\cdot 10^{-3}, 5\cdot 10^{-4}, 5\cdot 10^{-5}\}$, and a single-layer with $\{ 2^4, 2^6, 2^8, 2^{10} \}$ hidden nodes. 
The model resource costs $\gamma_1, \dotsc, \gamma_M$ are based on the model class and w.l.o.g. internally normalized, such that $\sum_{i=1}^M \gamma_i = 1$.
For the MLPs, they represent the number of hidden nodes, and for HTs, they refer to the maximal byte size (ie., 2KB, 4KB, ... 512KB, 1024KB).
All experiments are evaluated prequentially in a test-then-train manner, and a Python implementation is publicly available.
We have incorporated the implementation of \texttt{CAND} \cite{DBLP:conf/ijcnn/GunasekaraGPB22} into this framework, however, to enable the extension to different base learners, we deviate from their work by measuring performance using accuracy uniformly across all base learners, rather than relying on loss.
For ARF, SRP, and Shrubs, base learner HT, ensemble size $50$, grace period $50$, split confidence $0.01$, and $60\%$ as subspace size were chosen.
All experiments were run on a 10-core Broadwell processor (Intel 2630v4).

\subsection{Distribution of $X_i$}
\begin{figure}[t]
    \centering
     \subfigure[MLP, $k=12$.]{
        \includegraphics[width=0.75\linewidth]{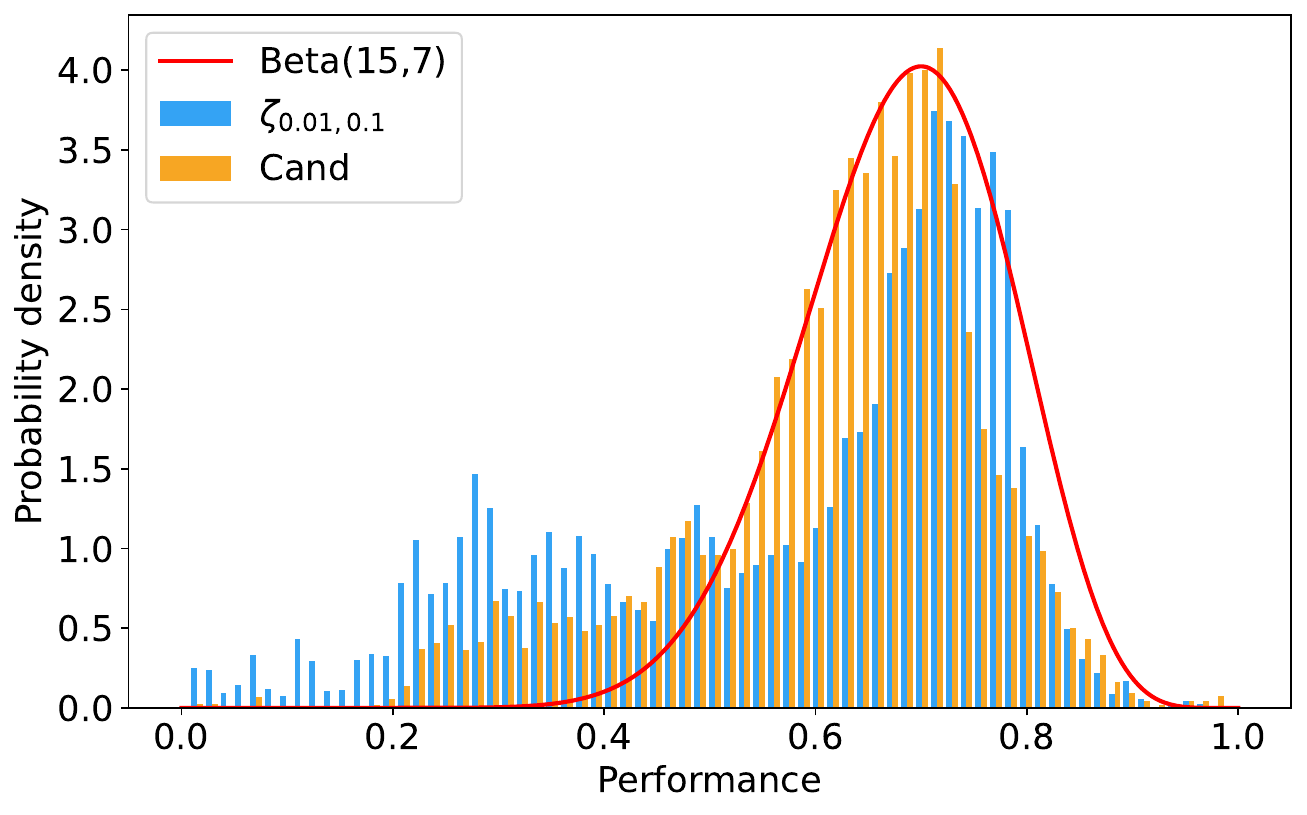}
        \label{subfig:dist_mlp}
    }
    \subfigure[HT, $k=8$.]{
        \includegraphics[width=0.75\linewidth]{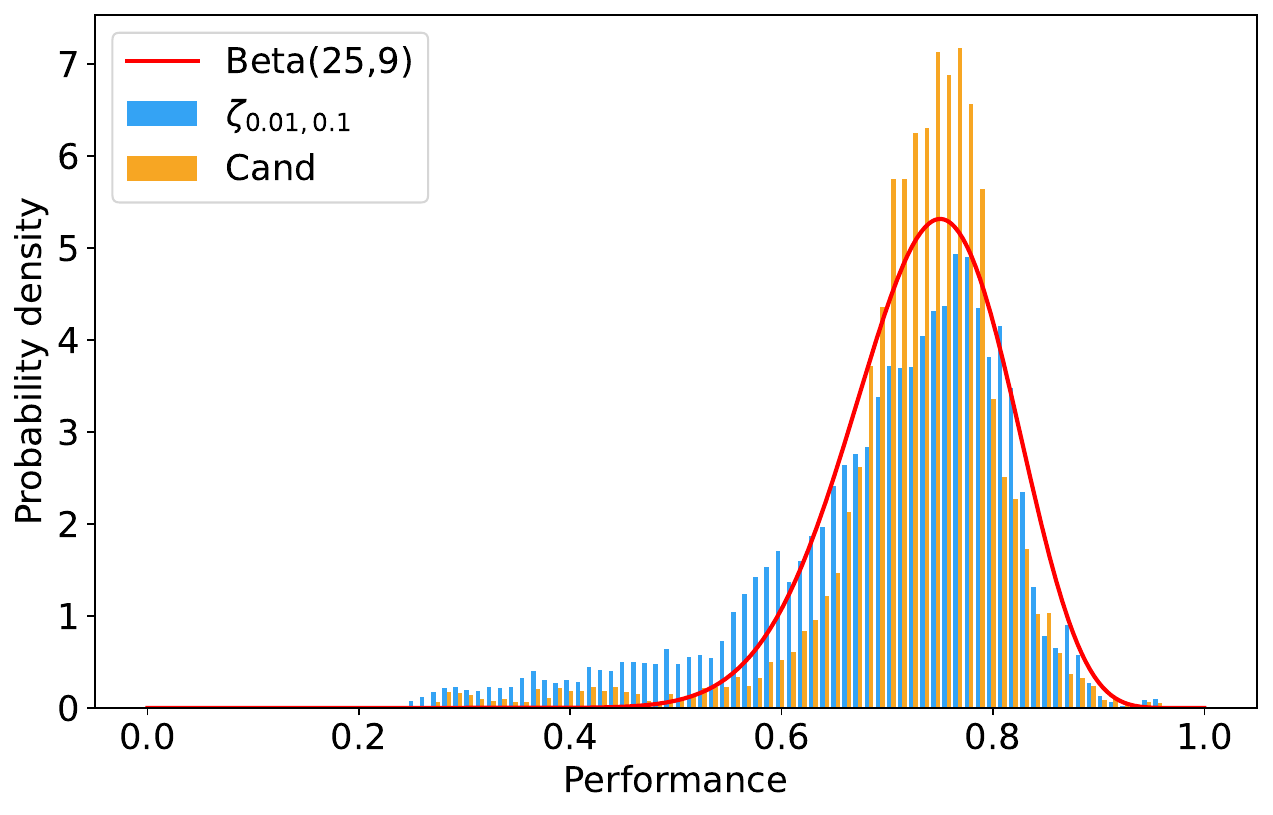}
        \label{subfig:dist_ht}
    }
        \caption{Probability density of $X_i=L(f_i)$ under $\zeta$-policy (with parameter $\zeta=0.01$ and $\epsilon=0.1$) and \texttt{cand}-policy, for data stream WISDM with different $k$, model pool size $M=20$ and base learner (\ref{subfig:dist_mlp}) MLP and (\ref{subfig:dist_ht}) Hoeffding tree. A Beta distribution Beta($\alpha, \beta$) is overlaid to visualize the approximation of the observed $X_i$.}
        \label{fig:beta_dist}
\end{figure}
In Section \ref{sec:theory}, we introduced a stochastic model to describe our setting during a training step. We assumed $L(X_i) = X_i \sim \text{Beta}(\alpha,\beta)$. In this experiment, we show that the Beta distribution is a good approximation of the distribution of $X_i$.
Figure \ref{fig:beta_dist} presents a histogram illustrating the performance values of all base learners in the pool $f$ over time for the policies $\zeta$ and \texttt{CAND}.
A Beta($\alpha, \beta$) distribution can be overlaid onto the distribution of all $X_i$ under $\zeta$ or \texttt{CAND} policy for the same $\alpha, \beta$ values. This demonstrates that Beta($\alpha, \beta$) serves as an effective approximation of $X_i=L(f_i)$, as proposed when introducing our stochastic model in Section \ref{subsec:stochastic_model}. Similar results are observed for different values of $k$ and base learners.

\subsection{HEROS with Various Policies}



\begin{table*}[t]

\caption{Summary with AUROC and total resource costs in kWh, $M=50$ and $k=30$, and base learner MLP for three random repetitions. The notation for the $\zeta$-policy is written as $\zeta_{\zeta \text{value}, \epsilon}$.}
\centering

\resizebox{\linewidth}{!}{

\begin{tabular}{l l | l l l l l l l l l l}

\toprule
Data stream & & Cand & Random & Perform-best & Perform-worst & Cheapest & Expensive & $\zeta_{0.01, 0.1}$ & $\zeta_{0.05, 0.1}$ & $\zeta_{0.01, 0.2}$ & $\zeta_{0.05, 0.2}$ \\
\midrule
airlines & Auroc & 0.7423& 0.7465& \textbf{0.7516}& 0.7245& 0.7456& 0.7504& 0.7513& 0.7508& 0.7510& 0.7503 \\
 & kWh & 2.4617& 3.1672& 2.8577& 2.1482& 1.2267& 3.7548& 3.1326& 2.8095& 3.1141& 2.4144 \\
electricity & Auroc & 0.9406& 0.9514& \textbf{0.9663}& 0.8653& 0.9624& 0.9650& 0.9662& 0.9662& 0.9630& 0.9626 \\
 & kWh & 0.0618& 0.0627& 0.0642& 0.0603& 0.0468& 0.0621& 0.0619& 0.0609& 0.0618& 0.0619 \\
wisdm & Auroc & 0.9037& 0.9173& 0.9370& 0.7663& \textbf{0.9485}& 0.9308& 0.9381& 0.9355& 0.9336& 0.9332 \\
 & kWh & 0.0128& 0.0127& 0.0121& 0.0122& 0.0085& 0.0131& 0.0127& 0.0127& 0.0127& 0.0127 \\
covtype & Auroc & 0.9831& 0.9859& \textbf{0.9904}& 0.9526& 0.9861& 0.9900& \textbf{0.9904}& 0.9901& 0.9896& 0.9896 \\
 & kWh & 1.6462& 1.6391& 1.6603& 1.5769& 1.0795& 1.8335& 1.6377& 1.6522& 1.6463& 1.4625 \\
nomao & Auroc & 0.9912& 0.9923& \textbf{0.9941}& 0.9858& 0.9921& 0.9913& 0.9936& 0.9930& 0.9938& 0.9935 \\
 & kWh & 0.1733& 0.1534& 0.1577& 0.1517& 0.1371& 0.1829& 0.1419& 0.1414& 0.1377& 0.1309 \\
AGR$_a$ & Auroc & 0.9829& 0.9846& 0.9860& 0.9638& 0.9863& 0.9848& 0.9864& 0.9860& \textbf{0.9865}& \textbf{0.9865} \\
 & kWh & 1.3593& 1.4861& 1.6449& 1.4457& 1.1194& 1.6902& 1.6015& 1.6853& 1.6954& 1.6277 \\
AGR$_g$ & Auroc & 0.9725& 0.9750& 0.9768& 0.9511& 0.9768& 0.9747& \textbf{0.9772}& 0.9768& 0.9769& \textbf{0.9772} \\
 & kWh & 1.6120& 1.6604& 1.7233& 1.5852& 1.2070& 1.9344& 1.8429& 1.7003& 1.7729& 1.7028 \\
RBF$_f$ & Auroc & 0.7934& 0.8298& 0.8696& 0.6063& 0.8428& \textbf{0.8697}& 0.8694& 0.8694& 0.8662& 0.8663 \\
 & kWh & 1.4567& 1.4372& 1.4479& 1.4297& 1.1193& 1.4964& 1.4377& 1.4159& 1.3847& 1.4060 \\
RBF$_m$ & Auroc & 0.9437& 0.9509& \textbf{0.9596}& 0.8708& 0.9516& \textbf{0.9596}& \textbf{0.9596}& 0.9595& 0.9589& 0.9589 \\
 & kWh & 1.3825& 1.3796& 1.4185& 1.3723& 1.0881& 1.4524& 1.4609& 1.4898& 1.5353& 1.5184 \\
LED$_a$ & Auroc & 0.9568& 0.9567& 0.9572& 0.9530& 0.9563& \textbf{0.9574}& 0.9572& \textbf{0.9574}& 0.9572& 0.9571 \\
 & kWh & 2.1826& 2.1458& 2.2192& 2.0198& 1.4549& 2.1588& 2.0552& 2.0587& 2.0236& 2.0299 \\
LED$_g$ & Auroc & 0.9549& 0.9552& 0.9556& 0.9511& 0.9546& \textbf{0.9559}& 0.9556& 0.9558& 0.9557& 0.9558 \\
 & kWh & 2.0205& 2.0138& 2.0853& 1.9881& 1.4648& 2.2914& 2.2723& 2.2681& 2.2531& 2.5509 \\
\midrule
\multicolumn{2}{l|}{Mean rank (AUROC)} & 8.73& 7.55& 2.73& 10.00& 6.55& 4.45& \textbf{2.64}& 3.64& 4.18& 4.55 \\
\multicolumn{2}{l|}{Mean rank (kWh)} & 5.82& 5.91& 7.00& 2.91& \textbf{1.09}& 9.09& 6.55& 5.73& 5.82& 5.09 \\

\bottomrule
\end{tabular}
}

\label{table:policies_nn}
\end{table*}
The results presented in Table \ref{table:policies_nn} compare the predictive performance in terms of AUROC and the resource consumption in kWh of HEROS with MLPs under different policies on 11 datasets. 
Evaluation results for base learner HT and different $k$ can be found in the Appendix. 
During the initial exploration, we were able to reproduce the results for \texttt{CAND} as given in \cite{DBLP:conf/ijcnn/GunasekaraGPB22}. Except for the dataset RBF$_f$, where we obtain a worse accuracy, likely due to a different hyperparameter configuration of its gradual drift. 
Hence, we are confident that our implementation matches the original results. However, for consistent evaluation, we will now report AUROC and use ensemble sizes of $50$.
As expected, the results confirm our assumption in Section \ref{sec:algorithm} that \texttt{perform-best} (and \texttt{perform-worse}) achieves almost the highest (and lowest) predictive performance and highest (and lowest) mean rank over the stream. We expected an analogous behavior for resource consumption and observe that \texttt{cheapest} (\texttt{expensive}) has the lowest (highest) average resource consumption across the stream.
\texttt{cand} has high AUROC values and ranks second worst overall with a mean of $8.73$, and its resource consumption lags behind our $\zeta$ policy.
As we have proven  stochastically in Section \ref{sec:theory}, and now show empirically $\zeta$-policy has higher AUROC than \texttt{CAND}, and $\zeta$-policy is slightly better than \texttt{perform-best}. 
The advantage of $\zeta_{0.01, 0.1}$ over \texttt{perform-best} may stem from its low $\zeta$ value, which  facilitates training a broader range of models with lower resource costs but still satisfactory predictive performance.
Moreover, analogously to Theorem \ref{theorem:perform-best-perf}, the predictive capabilities under $\zeta$-policy are dependent on $\zeta$, and for larger $\zeta$ values, the mean rank in terms of AUROC worsens. 
If $M$ and the $\zeta$ values are too small, the $1-\zeta$ range (Alg. \ref{alg:zeta-policy}) may be too small for selecting more resource-efficient models (cf. mean rank of $\zeta_{0.01, 0.1}$ and $\zeta_{0.01, 0.2}$ to \texttt{CAND} in Table  \ref{table:policies_nn}). 
Table \ref{table:policies_nn} also illustrates how to best choose $\epsilon$ and $\zeta$: for large $\epsilon$, the performance converge to \texttt{random}, since \texttt{random} equals $\zeta_{.,1}$. As $\zeta$ approaches $0$, the performance aligns with \texttt{perform-best}; at $\zeta=1$ corresponds to \texttt{cheapest}.
Nevertheless, HEROS is a robust streaming hyperparameter optimization method, showing low sensitivity to small $\zeta$ and $\epsilon$ settings.
However, the results also confirm our theory, the lower average resource consumption under $\zeta$ policy, from Theorem \ref{theorem:ressource_cand} and Theorem \ref{theorem:ressource_performbest}.
Figure \ref{fig:trade-off} presents the result for the post-hoc Holm-adjusted Wilcoxon test with $95\%$ confidence level. We note that in terms of predictive performance, no significant difference can be found  among \texttt{perform-best}, \texttt{expensive}, and $\zeta$-policy.

\begin{figure}[t]
    \centering
    \subfigure[HT]{
        \includegraphics[width=0.45\linewidth]{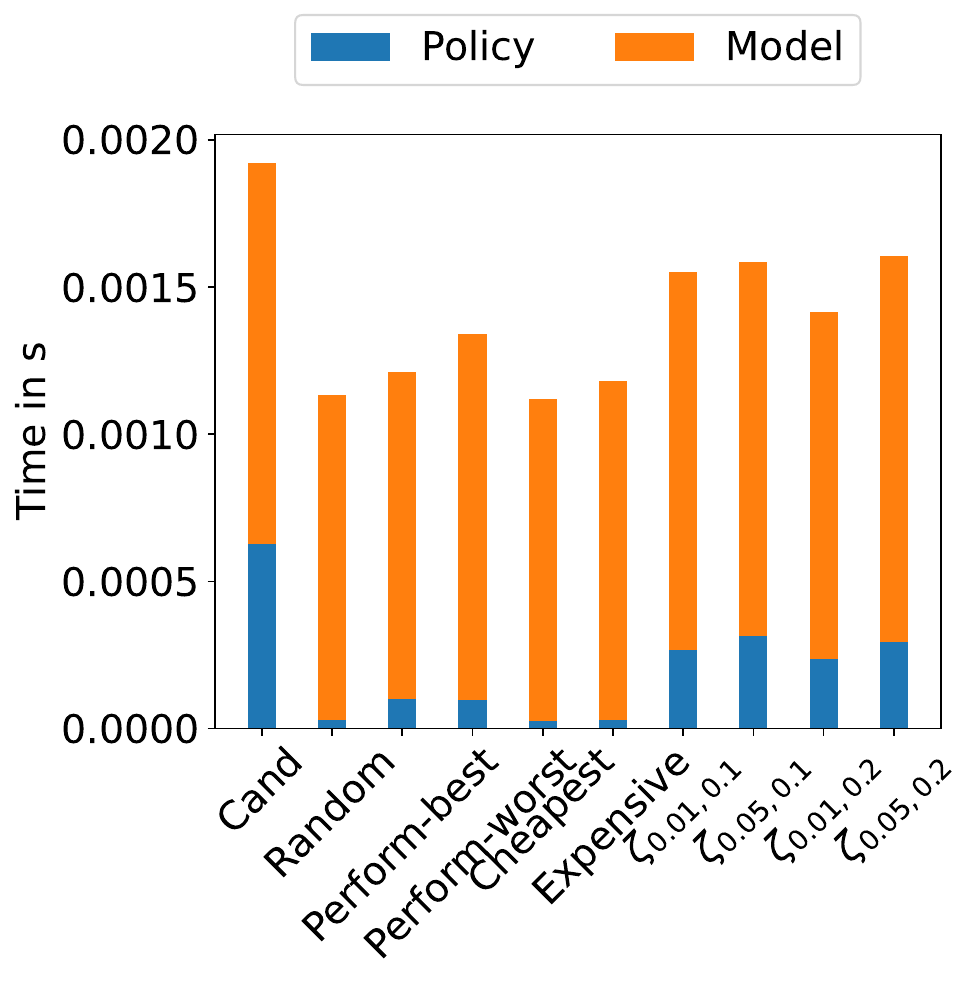}
        \label{subfig:resource_distribution_ht}
    }
    \subfigure[MLP]{
        \includegraphics[width=0.45\linewidth]{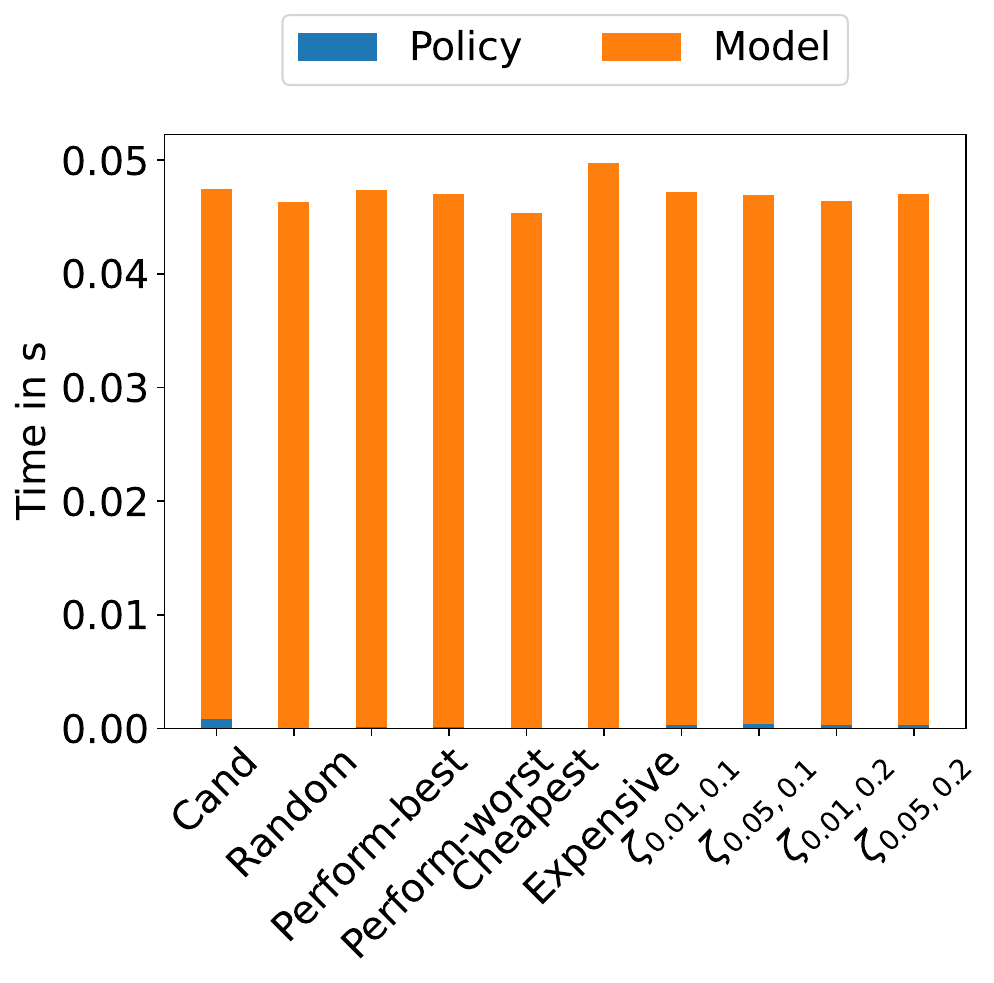}
        \label{subfig:resource_distribution_mlp}
    }
    \caption{Distribution of average time (in s) used for model evaluation and policy execution per instance for data stream electricity, $P=50$ and $k=30$ for base learners HT and MLP. (\ref{subfig:resource_distribution_ht}) For \texttt{cheapest} and \texttt{expensive}, policy computation accounts for only $2-3\%$ of total costs, keeping overall resource use low due to inexpensive models. In contrast, for \texttt{Cand}, policy computation makes up over $32\%$ , and $\zeta$ policy under $18\%$, leading to higher total resource use. (\ref{subfig:resource_distribution_mlp}) With MLP base learner, policy computation is small (around $1\%$), allowing for greater resource savings.}
    \label{fig:resources_various_base_learner}
\end{figure}
The predictive performance under $\zeta$ policy with base learner HT in the Appendix is delivering consistently the best results overall. However, the resource consumption differs. Both the energy usage in kWh and the mean rank are relatively close across policies. An analysis of the measured energy distribution reveals that HT performs prediction and training very quickly, meaning most of the computational costs are incurred during policy calculation (cf. Fig. \ref{fig:resources_various_base_learner}). 

As model complexity increases, even lightweight models may offer a resource advantage using HEROS. However, the energy measurements for the results with HT as base learner (given in the Appendix) might also be biased due to the software architecture: HEROS and its policies are implemented in CapyMOA using Python. Many base learners (such as HT) in CapyMOA are, however, implemented in Java utilizing its Java interface. Hence, there is an imbalance in measuring the energy consumption of our policies and some base learners like HT. Since MLPs are implemented in Python, this leads to more consistent results.

\subsection{Drift Adaption}
\begin{figure}[t]
    \centering
    \subfigure[Emission (CO$_2$)]{
        \includegraphics[width=0.85\linewidth]{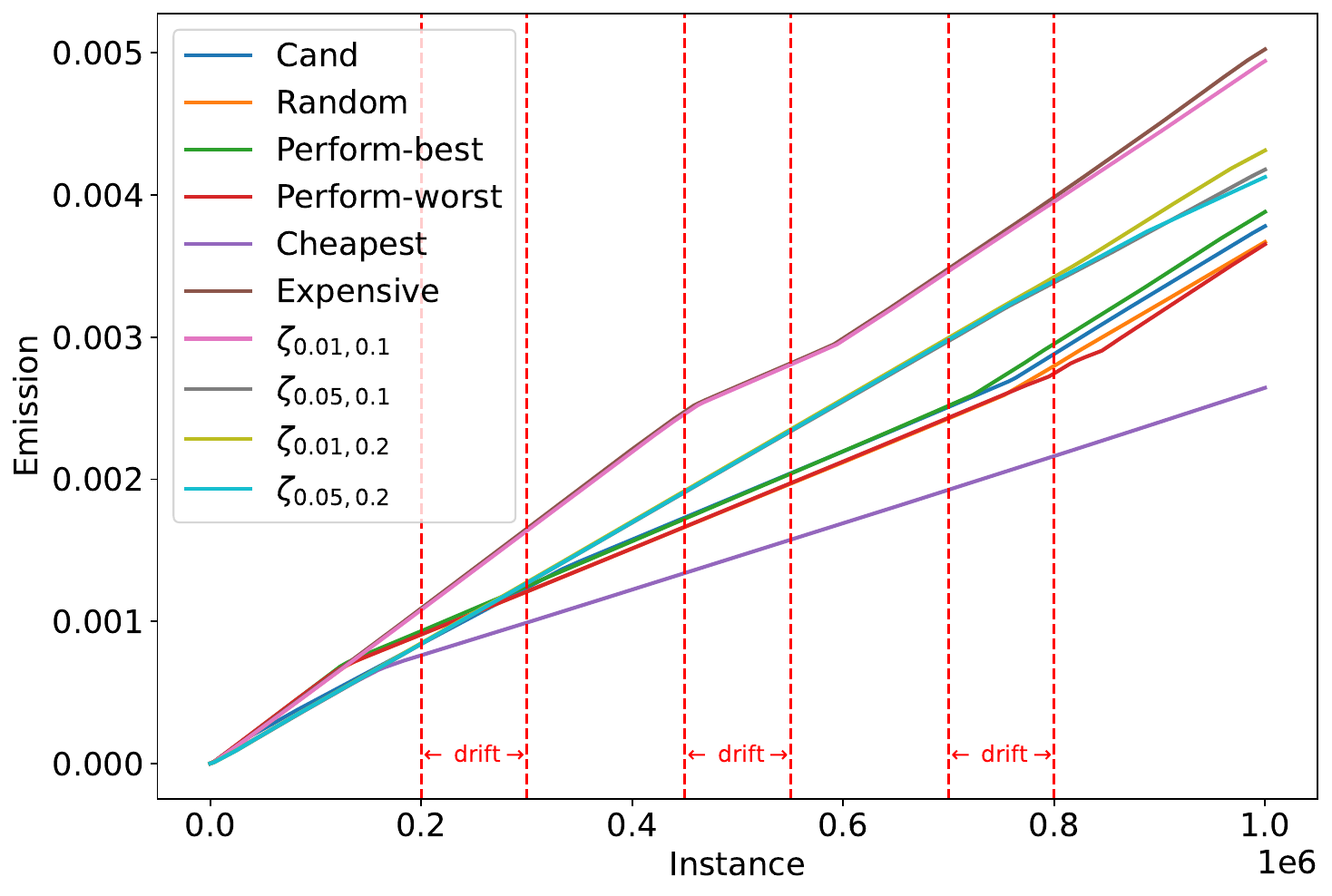}
        \label{subfig:emission-agr-g}
    }
    \subfigure[Resource cost]{
        \includegraphics[width=0.85\linewidth]{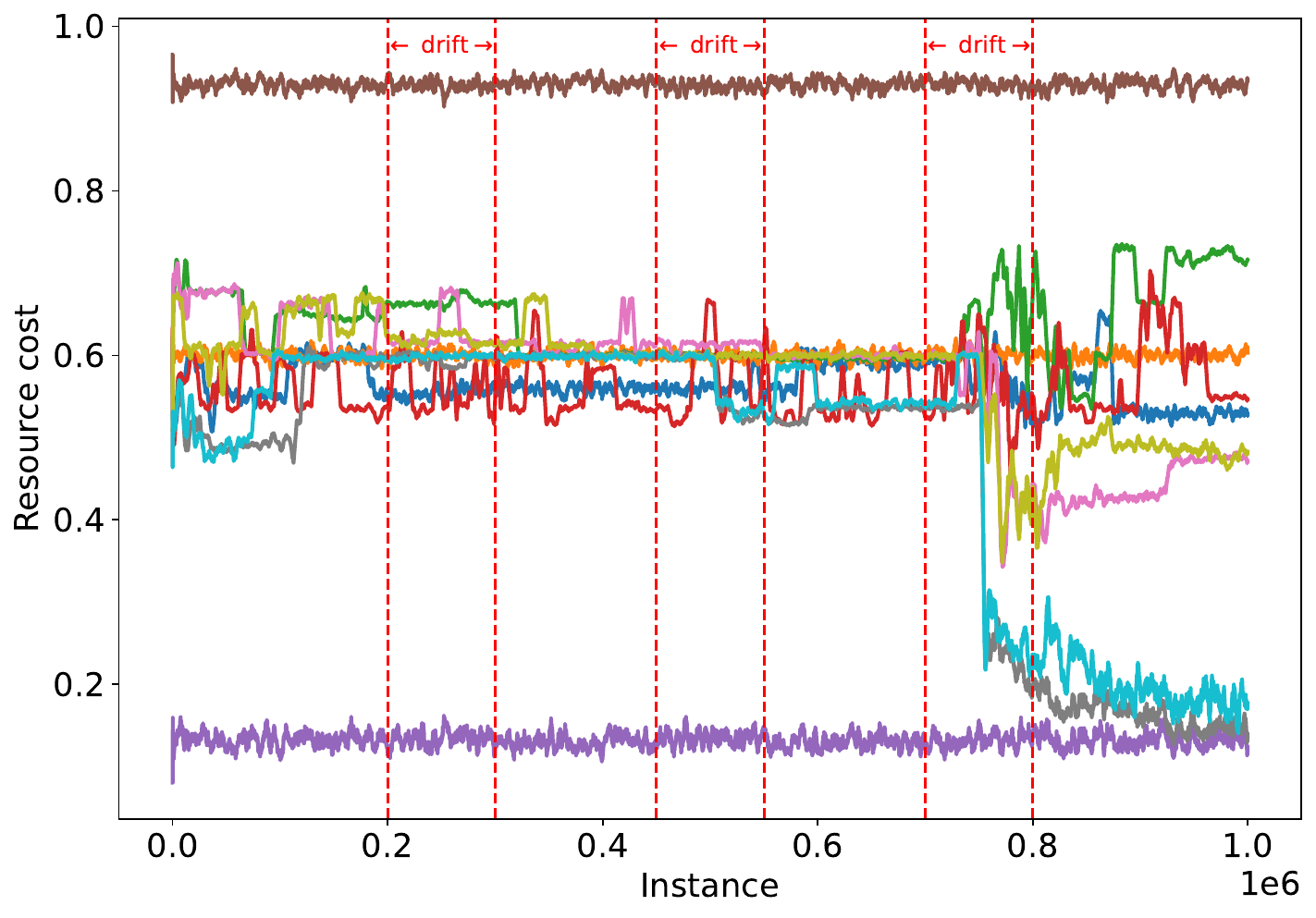}
        \label{subfig:res-agr-g}
    }
    \caption{Resource consumption per training step with $M=50$, $k=30$ for base learner MLP on data stream AGR$_g$. Gradual drift is marked with vertical lines and width $\leftarrow$ drift $\rightarrow$. Resources are smoothed with a sliding window of size $300$.}
     \label{fig:resource_gradual_drift}
\end{figure}
To further investigate the dynamic properties of our framework, we now highlight the per instance resource consumption over time during prequential evaluation for all introduced policies in Figure \ref{fig:resource_gradual_drift}. AGR$_g$ contains four concepts with 3 gradual drifts marked with vertical lines in the figure.
As noted in the previous section, \texttt{expensive} and \texttt{cheapest} have the highest and lowest, respectively, resource costs highlighting the range of resource costs.
In reaction to a gradual drift, the resource consumption under $\zeta$-policy changes, which implies that another subset of $k$ models is selected for training, as these models might better adapt to the new concept.
Particularly during the gradual concept change, the selected subset and their resource consumption vary greatly under the $\zeta$-policy.
The results confirm that a heterogeneous model pool mitigates the problem of  hyperparameter tuning, with better-suited models used during or after drift.
Similar observations can be made for abrupt drifts given in the Appendix.

\subsection{HEROS against Ensemble Learners}
\begin{table}[t]

\caption{Auroc for state-of-the-art ensembles for 3 random repetitions. Results with HEROS using $\zeta$-policy with $\zeta=0.05$ and $\epsilon=0.2$ are with base learner MLP, $P=50$ and $k=30$.}
\centering

\begin{tabular}{lllll}
\toprule
Data stream & ARF & SRP & Shrubs & HEROS \\
\midrule
airlines & 0.681 & 0.737 & 0.584 & \textbf{0.750} \\
electricity & 0.961 & 0.957 & 0.932 & \textbf{0.963} \\
wisdm & \textbf{0.947} & 0.943 & 0.873 & 0.933 \\
covtype & \textbf{0.994} & 0.993 & 0.921 & 0.990 \\
nomao & 0.993 & \textbf{0.994} & 0.989 & 0.994 \\
AGR$_a$ & 0.940 & 0.971 & 0.859 & \textbf{0.987} \\
AGR$_g$ & 0.921 & 0.959 & 0.841 & \textbf{0.977} \\
RBF$_f$ & \textbf{0.935} & 0.927 & 0.739 & 0.866 \\
RBF$_m$ & \textbf{0.977} & 0.968 & 0.737 & 0.959 \\
LED$_a$ & \textbf{0.958} & 0.957 & 0.867 & 0.957 \\
LED$_g$ & 0.955 & 0.955 & 0.860 & \textbf{0.956} \\
\midrule
Mean Rank & \textbf{1.909} & 2.182 & 4.000 & \textbf{1.909} \\
 \bottomrule
\end{tabular}

\label{tab:ensembles}
\end{table}

Finally, we compare the current state-of-the-art streaming algorithms ARF \cite{DBLP:journals/ml/GomesBRBEPHA17}, SRP \cite{DBLP:conf/icdm/GomesRB19} and Shrubs \cite{DBLP:conf/aaai/BuschjagerHM22}, which do not consider costs, to HEROS using the $\zeta$-policy in Table \ref{tab:ensembles}.
We were able to reproduce the results for ARF and SRP as given by the authors \cite{DBLP:conf/icdm/GomesRB19}.
The results on the synthetic data with induced drift clearly demonstrate that our HEROS framework achieves a superior predictive performance compared to SRP and Shrubs, and is competitive with ARF. 
Avoiding hyperparameter tuning by providing multiple trained models of different configurations allows HEROS to react better to drifts than current state-of-the-art ensemble methods.

\section{Conclusion}
\label{sec:conclusion}
In this work, we address the challenge of maintaining a heterogeneous ensemble under resource constraints on an incoming and evolving data stream. We introduced an MDP to unify online learning and support different policies for training HEROS, including our novel $\zeta$-policy.
In comparison to \texttt{CAND}, we have stochastically proven that our proposed policy achieves higher average predictive performance for small $\zeta$, while the average resource costs in terms of kWh are lower.
Moreover, the average performance is asymptotically at most $\zeta$ worse than applying the \texttt{perform-best} policy.
Our experiments endorse our theoretical results and also demonstrate the predictive performance of HEROS to be better on drifting data streams.
On the basis of the theoretical and practical results, one can view the framework as a green and sustainable approach to online learning.
It delivers impressive resource savings during training,  while maintaining good predictive performance and enabling users to control $\zeta$ and thus balance the trade-off between quality and environmental costs.

\section*{Acknowledgments}
The research in this paper was supported by the ``TOPML: Trading Off Non-Functional Properties of Machine Learning'' project funded by Carl Zeiss Foundation, grant number P2021-02-014.
Part of this research has been funded by the Federal Ministry of Research, Technology and Space of Germany and the state of North Rhine-Westphalia as part of the Lamarr Institute for Machine Learning and Artificial Intelligence.

\bibliographystyle{IEEEtran}
\bibliography{IEEEabrv,./bibliography}

\vspace{11pt}

{\appendices
\section{Supplementary Theory}
\begin{proof}[Proof of Lemma \ref{lemma:cand}]
  First, we observe
    \begin{align*}
    \lim_{M\to\infty} \mathbb{P}(\mbox{a model was selected twice})= 0.
\end{align*}
  Conditional on this event, we first consider $X^C_k$. 
For $M$ large and $k$ large, the arithmetic mean of $X_{(M)}, \dotsc, X_{(M-\lfloor\frac{k}{2}\rfloor)}$ converges in probability to $1$. This is since the support of $X_{(M)}, \dotsc, X_{(M-\lfloor\frac{k}{2}\rfloor)}$ is $[0,1]$ and for $i\in \{1, \dotsc, \lfloor\frac{k}{2}\rfloor\}$ is $P(X_{(M-i-1)} > 1-\hat{\epsilon}) > 0$ by construction, which means that the values of $X_{(M-i-1)}$ are near $1$ no matter how small $\hat{\epsilon}>0$ is. Therefore, when $M \to \infty$ the probability of observing at least one $X_{(M-i-1)}$ in the range $(1-\hat{\epsilon}, 1]$ approaches $1$.
The arithmetic mean of the random selection $X_{1,r}^C, \dotsc, X_{\lceil\frac{k}{2}\rceil,r}^C$ is $\mathbb{E}(X)$ as they are selected randomly.
By the strong law of large numbers 
   \begin{align*}
       & \lim_{k\to \infty} {\frac{2}{k}} \sum_{i=1}^{k/2} X_{(M-i-1)} + {\frac{2}{k}} \sum_{i=1}^{k/2} X_{i,r} = 1 + \mathbb{E}(X) \\
       \Leftrightarrow & \lim_{k\to \infty} {\frac{2}{k}} ( \sum X_{(M-i-1)} + \sum X_{i,r} ) = 1 + \mathbb{E}(X) \\
   \Leftrightarrow & \lim_{k\to \infty} \frac{1}{k} X^{C} = \frac{1}{2} + \frac{1}{2} \mathbb{E}(X)
   \end{align*}
   The first part of the claim follows.

    As with high probability, no model is selected twice, and with high probability $\gamma^C$ has the same distribution as the sum of $2 \cdot \frac{k}{2} $ independent random variables with the same distribution as $\gamma$. Therefore, the strong law of large numbers yields
    \begin{align*}
        \lim_{k\to \infty} \frac{1}{k}\gamma^{C} = \mathbb{E}(\gamma)
    \end{align*}
    almost surely.    
\end{proof}

\begin{proof}[Proof of Lemma \ref{lemma:zeta}]
    As the $X_i^{\zeta}$ and $\gamma_i^{\zeta}$ are independent for all $i \in \{1, \dotsc, k\}$, the $X_i^{\zeta}$ are then identically distributed and independent with distribution $\mathbb{P}(X\in \cdot \,\vert X \geq 1-\zeta)$. We immediately get the worst-case lower bound by selecting $k$ models from the outermost edge $(1-\zeta)\cdot X_{(M)}$:
     $\sum X_i^{\zeta}\geq k(1-\zeta)$. By the strong law of large numbers,
    \begin{align*}
    \lim_{k\to\infty}\frac{1}{k} \sum X_i^{\zeta} = \mathbb{E}(X\vert X \geq 1-\zeta) \quad \mbox{almost surely}. 
    \end{align*}
    By definition, the $\gamma_1^{\zeta}, \gamma_2^{\zeta}, \dotsc, \gamma_k^{\zeta}$ have the smallest resource values in $\{X \geq 1-\zeta\}$. Therefore, 
    $$
    \max_{i\leq k} \gamma_i^{\zeta}\to 0,\quad \mbox{almost surely},
    $$
  as $M\to\infty$.
\end{proof}

\begin{proof}[Proof of Theorem \ref{theorem:performance-cand}]
    From Lemma \ref{lemma:cand} and Lemma \ref{lemma:zeta} we know, the arithmetic mean converges in probability under $\zeta$-policy to the conditional expected value $\mathbb{E}(X\vert X \geq 1-\zeta)$ and under \texttt{CAND} policy to $\frac{1}{2} + \frac{1}{2} \mathbb{E}(X)$.
   As $\mathbb{E}(X\vert X \geq 1-\zeta)\geq 1-\zeta$ is always true, we can estimate the following:
\begin{align}
   & \mathbb{E}(X\vert X \geq 1-\zeta) > \frac{1}{2} + \frac{1}{2} \mathbb{E}(X) \\
   \Leftrightarrow & 1 - \zeta > \frac{1}{2} + \frac{1}{2} \mathbb{E}(X) \\
   \Leftrightarrow & \frac{1}{2} (1-\mathbb{E}(X)) > \zeta.
\end{align}
If $\frac{1}{2} (1-\mathbb{E}(X)) > \zeta$, then the first part of the claim follows.
    For $X$ Beta($\alpha, \beta$) distributed the expectation is $\mathbb{E}(X) = \frac{\alpha}{\alpha + \beta}$ and the last part of the claim also follows.
\end{proof}

\begin{proof}[Proof of Theorem \ref{theorem:ressource_cand}]
   With Lemma \ref{lemma:cand} and Lemma \ref{lemma:zeta} we confirm that
$    
        \bar{\gamma}^{\zeta} < \mathbb{E}(\gamma) = \bar{\gamma}^{C}
$    
    is true.
\end{proof}

\begin{lemma}\label{lemma:performbest}
    The asymptotic behavior of the average of $X^{pb}$ and $\gamma^{pb}$ under \texttt{perform-best} policy converges in probability to $1$ and $\mathbb{E}(\gamma)$, respectively.
\end{lemma}
\begin{proof}[Proof of Lemma \ref{lemma:performbest}]
    Since $M$ tends to infinity first, $\frac{1}{k}X^{pb}$ converges almost surely to $1$. 
    The determination of the average resource costs follows directly from Theorem \ref{theorem:ressource_cand} as resources are not considered in the selection process in the \texttt{perform-best} policy and $X$ and $\gamma$ are independent. Therefore, the average expected resource consumption of $\frac{1}{k}\gamma^{pb}$ for $M\to \infty$ and then $k\to\infty$ converges to $\mathbb{E}(\gamma)$.
\end{proof}
\begin{proof}[Proof of Theorem \ref{theorem:perform-best-perf}]
    The difference of the overall performances $X^{pb}$ and $X^{\zeta}$ in their asymptotic behavior analyzed in Lemma \ref{lemma:performbest} and \ref{lemma:zeta} can be estimated as
    \begin{align*}
        \lim_{k\to\infty} \lim_{M\to\infty} \frac{1}{k} |X^{pb} - X^\zeta| \leq  \zeta.
    \end{align*}
\end{proof}
\begin{proof}[Proof of Theorem \ref{theorem:ressource_performbest}]
    Analogous to the proof of Theorem~\ref{theorem:ressource_cand}.
\end{proof}

\section{Supplementary Experiments}

\begin{table*}[t]
\caption{Summary with AUROC and total resource cost in kWh, $P=50$ and $k=20$, and base learner Hoeffding tree for three random repetitions (seeds 1, 2, 3). The notation for the $\zeta$-policy is written as $\zeta_{\zeta \text{value}, \epsilon}$.}
\centering

\resizebox{\linewidth}{!}{

\begin{tabular}{l l | l l l l l l l l l l}
\toprule
Data stream & & Cand & Random & Perform-best & Perform-worst & Cheapest & Expensive & $\zeta_{0.01, 0.1}$ & $\zeta_{0.05, 0.1}$ & $\zeta_{0.01, 0.2}$ & $\zeta_{0.05, 0.2}$ \\
\midrule
airlines & AUROC & 0.6503& 0.6497& 0.6719& 0.6497& 0.6717& 0.6717& 0.6722& \textbf{0.6739}& 0.6709& 0.6696 \\
 & kWh & 0.0300& 0.0380& 0.0229& 0.0243& 0.0446& 0.0372& 0.0275& 0.0314& 0.0308& 0.0314 \\
electricity & AUROC & 0.8402& 0.8564& 0.8861& 0.8056& 0.8677& 0.8720& 0.8896& \textbf{0.8923}& 0.8889& 0.8850 \\
 & kWh & 0.0043& 0.0028& 0.0018& 0.0017& 0.0019& 0.0017& 0.0019& 0.0022& 0.0024& 0.0023 \\
wisdm & AUROC & 0.8849& 0.8944& 0.9147& 0.8472& 0.8760& 0.8709& 0.9132& 0.9006& \textbf{0.9149}& 0.9133 \\
 & kWh & 0.0006& 0.0005& 0.0005& 0.0006& 0.0005& 0.0005& 0.0006& 0.0006& 0.0005& 0.0006 \\
covtype & AUROC & 0.9570& 0.9622& 0.9752& 0.9411& 0.9706& 0.9707& \textbf{0.9756}& 0.9752& 0.9740& 0.9741 \\
 & kWh & 0.0399& 0.0353& 0.0370& 0.0327& 0.0352& 0.0365& 0.0451& 0.0417& 0.0324& 0.0350 \\
nomao & AUROC & 0.9471& 0.9503& 0.9580& 0.9284& 0.9441& 0.9397& 0.9580& 0.9497& \textbf{0.9601}& 0.9510 \\
 & kWh & 0.0042& 0.0035& 0.0032& 0.0033& 0.0033& 0.0046& 0.0035& 0.0031& 0.0033& 0.0033 \\
AGR$_a$ & AUROC & 0.9063& 0.8955& 0.9461& 0.8778& 0.8760& 0.8703& 0.9465& 0.9447& \textbf{0.9597}& 0.9458 \\
 & kWh & 0.1199& 0.1083& 0.1111& 0.1399& 0.1180& 0.1354& 0.1273& 0.2046& 0.2133& 0.1268 \\
AGR$_g$ & AUROC & 0.8784& 0.8702& 0.9160& 0.8523& 0.8496& 0.8434& 0.9221& 0.9219& \textbf{0.9309}& 0.9184 \\
 & kWh & 0.1250& 0.1767& 0.1451& 0.1039& 0.1040& 0.1061& 0.1241& 0.1175& 0.1231& 0.1526 \\
RBF$_f$ & AUROC & 0.5955& 0.6128& 0.6734& 0.5674& 0.6253& 0.6242& 0.6743& \textbf{0.6760}& 0.6668& 0.6708 \\
 & kWh & 0.0764& 0.0540& 0.0549& 0.0769& 0.0727& 0.0752& 0.0884& 0.0843& 0.1095& 0.0931 \\
RBF$_m$ & AUROC & 0.7691& 0.7871& 0.8570& 0.7520& 0.8058& 0.8050& 0.8556& \textbf{0.8599}& 0.8510& 0.8521 \\
 & kWh & 0.0520& 0.0468& 0.0520& 0.0463& 0.0570& 0.0526& 0.0578& 0.0806& 0.0616& 0.0639 \\
LED$_a$ & AUROC & 0.9423& 0.9412& 0.9492& 0.9500& 0.9501& 0.9505& 0.9495& 0.9509& 0.9507& \textbf{0.9520} \\
 & kWh & 0.0599& 0.0809& 0.0896& 0.1205& 0.1113& 0.0705& 0.1007& 0.0762& 0.0622& 0.0782 \\
LED$_g$ & AUROC & 0.9406& 0.9405& 0.9457& 0.9457& \textbf{0.9488}& 0.9486& 0.9441& 0.9471& 0.9483& 0.9485 \\
 & kWh & 0.0867& 0.0616& 0.0618& 0.0925& 0.0909& 0.0908& 0.0957& 0.0715& 0.0708& 0.0804 \\
\midrule
\multicolumn{2}{l|}{Mean rank (AUROC)} & 8.00& 7.82& 3.91& 8.91& 6.36& 6.82& 3.18& \textbf{2.91}& 3.18& 3.91 \\
\multicolumn{2}{l|}{Mean rank (kWh)} & 6.00& 5.00& \textbf{3.64}& 5.18& 4.91& 5.18& 7.00& 6.00& 5.64& 6.45 \\
\bottomrule
\end{tabular}

}

\label{tab:hoeffding_tree}
\end{table*}

In Table \ref{tab:hoeffding_tree}, the remaining results for base learner HT are presented.
Figure \ref{fig:resource_abrupt_drift} shows the resource consumption per training step for HEROS with MLP for the base learner, analyzing an abrupt drift. 
\begin{figure}[H]
    \centering
    \subfigure[Emission (CO$_2$)]{
        \includegraphics[width=0.85\linewidth]{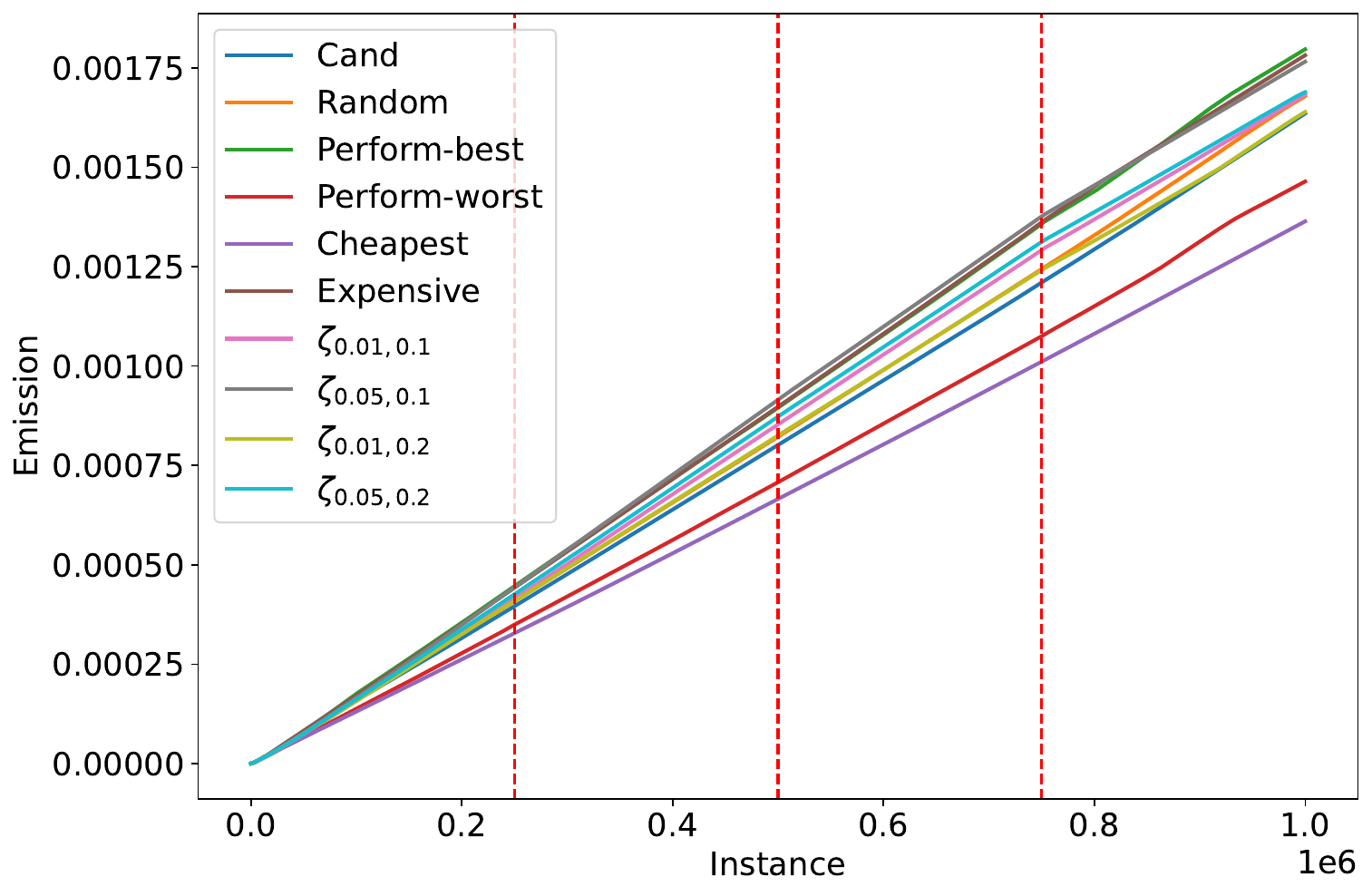}
        \label{subfig:emission-agr-a}
    }
    \subfigure[Resource cost]{
        \includegraphics[width=0.85\linewidth]{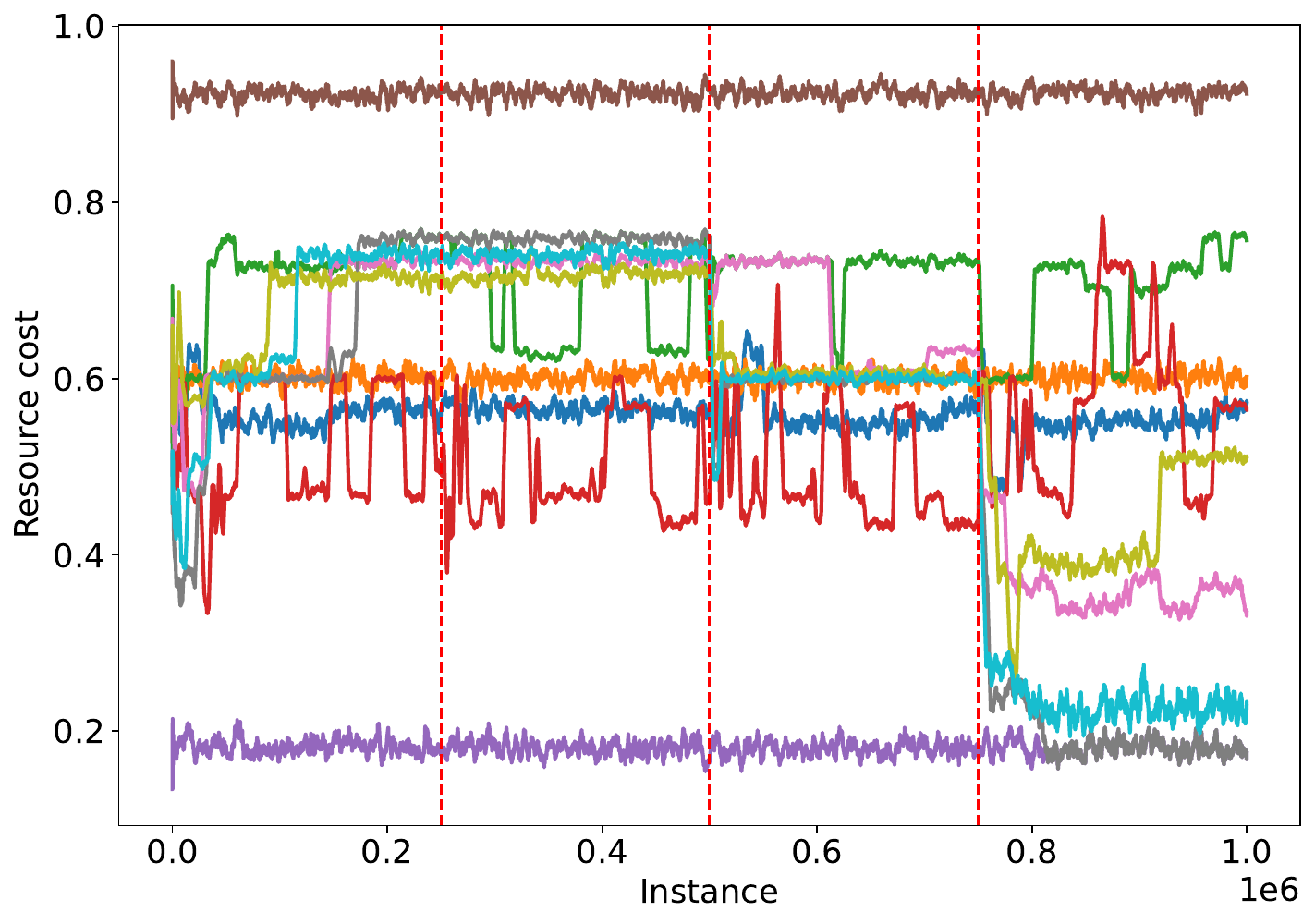}
        \label{subfig:res-agr-a}
    }
    \caption{Resource consumption per training step with $M=20$, $k=12$ for base learner MLP on data stream AGR$_a$. Abrupt drift is marked with vertical lines. Resources are smoothed with a sliding window of size $300$.}
     \label{fig:resource_abrupt_drift}
\end{figure}
}

\vfill

\end{document}